\documentclass[sigconf]{acmart}

\usepackage{soul}

\usepackage{multirow}
\usepackage{enumitem}
\usepackage{microtype}
\usepackage{comment}
\usepackage{hyperref}

\usepackage{subcaption}

\newcommand\BibTeX{B\textsc{ib}\TeX}
\newcommand{\stitle}[1]{\noindent{\textbf{#1}}}

\DeclareMathAlphabet{\mathpzc}{OT1}{pzc}{m}{it}

\usepackage[ruled,vlined]{algorithm2e}
\SetKwInput{KwInput}{Input}                
\SetKwInput{KwOutput}{Output}              
\SetKwInOut{Input}{Input}
\SetKwInOut{Output}{Output}
\SetKwInput{Parameter}{Parameter}
\SetKwInput{Parameters}{Parameters}
\SetKwInput{Algorithm}{Algorithm}
\SetKwInput{Remark}{Remark}
\SetKwInput{Variable}{Local variable}
\SetKwInput{Variables}{Local variables}
\SetKwInput{Function}{Local function}
\SetKwInput{Functions}{Local functions}
\SetKw{Let}{Let}
\SetKw{Add}{Add}
\SetKwComment{Comment}{{// }}{}

\usepackage{latexsym}
\usepackage{svrsymbols}
\usepackage{graphicx}
\usepackage{amsmath}

\definecolor{fgreen}{rgb}{0.0, 0.5, 0.0}

\newcommand{\myNum}[1]{(\emph{#1})}

\newcommand{\one} {\mathpzc{1} }
\newcommand{\two} {\mathpzc{2} }
\newcommand{\four} {\mathpzc{4} }

\newcommand{\bigS} {\mathcal{S} }
\newcommand{\bigX} {\mathcal{X} }
\newcommand{\bigY} {\mathcal{Y} }
\newcommand{\bigJ} {J}
\newcommand{\bigK} {K}
\newcommand{\bigU} {\mathcal{U}}
\newcommand{\bigH} {\mathcal{H} }
\newcommand{\bigD} {\mathcal{D} }
\newcommand{\bigC} {\mathcal{C} }
\newcommand{\bigA} {\mathcal{A} }
\newcommand{\bigG} {\mathcal{G} }

\newcommand{\smalld} {\mathpzc{dim} }
\newcommand{\lstmd} {\mathpzc{d} }
\newcommand{\smallh} {h}
\newcommand{\smalli} {\mathpzc{i} }
\newcommand{\smallx} {x}
\newcommand{\smally} {y}
\newcommand{\smalls} {s}
\newcommand{\smalll} {l}
\newcommand{\smallz} {z}
\newcommand{\smalln} {n}
\newcommand{\smallj} {j}
\newcommand{\smallk} {k}
\newcommand{\smallu} {u}
\newcommand{\smalla} {a}
\newcommand{\smallt} {t}
\newcommand{\smallv} {v}
\newcommand{\smallb} {b}
\newcommand{\smallm} {m}
\newcommand{\smallq} {q}

\newcommand{\smallr} {r}
\newcommand{\smallc} {c}
\newcommand{\smallf} {f}
\newcommand{\smalle} {e}
\newcommand{\smallp} {p}

\newcommand{\realR} {\mathbb{R} }

\newcommand{\noiob} {$\mathpzc{w/o}~\mathtt{IOB}$}

\newcommand{\ourmodel}{$\mathsf{LEONA}$}
\newcommand{\myvalue}[1] {$\mathtt{#1}$}
\newcommand{\subvalue}[1] {\mathtt{#1}}
\newcommand{\myspecial}[1] {\texttt{#1}}

\AtBeginDocument{%
  \providecommand\BibTeX{{%
    \normalfont B\kern-0.5em{\scshape i\kern-0.25em b}\kern-0.8em\TeX}}}

\begin{document}

\title[Linguistically-Enriched and Context-Aware Zero-shot Slot Filling]{Linguistically-Enriched and Context-Aware\\Zero-shot Slot Filling}

\author{A. B. Siddique}
\affiliation{%
  \institution{University of California, Riverside}
}
\email{msidd005@ucr.edu}

\author{Fuad Jamour}
\affiliation{%
  \institution{University of California, Riverside}
}
\email{fuadj@ucr.edu}

\author{Vagelis Hristidis}
\affiliation{%
  \institution{University of California, Riverside}
}
\email{vagelis@cs.ucr.edu}
\renewcommand{\shortauthors}{Siddique et al.}

\begin{abstract}
Slot filling is identifying contiguous spans of words in an utterance that correspond to certain parameters (i.e., slots) of a user request/query.
Slot filling is one of the most important challenges in modern task-oriented dialog systems. Supervised learning approaches have proven effective at tackling this challenge, but they need a significant amount of labeled training data in a given domain. However, new domains (i.e., unseen in training) may emerge after deployment. Thus, it is imperative that these models seamlessly adapt and fill slots from both seen and unseen domains -- unseen domains contain unseen slot types with no training data, and even seen slots in unseen domains are typically presented in different contexts.
This setting is commonly referred to as zero-shot slot filling.
Little work has focused on this setting, with limited experimental evaluation.
Existing models that mainly rely on context-independent embedding-based similarity measures fail to detect slot values in unseen domains or do so only partially. 
We propose a new zero-shot slot filling neural model, {\ourmodel}, which works in three steps. 
Step one acquires domain-oblivious, context-aware representations of the utterance word by exploiting (a) linguistic features such as part-of-speech; (b) named entity recognition cues; and (c) contextual embeddings from pre-trained language models. 
Step two fine-tunes these rich representations and produces slot-independent tags for each word. 
Step three exploits generalizable context-aware utterance-slot similarity features at the word level, uses slot-independent tags, and contextualizes them to produce slot-specific predictions for each word.
Our thorough evaluation on four diverse public datasets demonstrates that our approach consistently outperforms the state-of-the-art models by $17.52\%$, $22.15\%$, $17.42\%$, and $17.95\%$ on average for unseen domains on \myspecial{SNIPS}, \myspecial{ATIS}, \myspecial{MultiWOZ}, and \myspecial{SGD} datasets, respectively.

\end{abstract}

\maketitle

\section{Introduction}
\label{intro}
Goal-oriented dialog systems allow users to accomplish tasks, such as reserving a table at a restaurant, through an intuitive natural language interface (e.g., Amazon Alexa).
For instance, a user may issue the following utterance: \emph{``I would like to book a table at 8 Immortals Restaurant in San Francisco  for 5:30 pm today for 6 people''}.
For dialog systems to fulfill such a request, they first need to extract the parameter (a.k.a. slot) values of the request.
Slots in the restaurant booking domain include \myvalue{restaurant\_name} and \myvalue{city}, whose values in our example utterance are ``\myvalue{8} \myvalue{Immortals} \myvalue{Restaurant}'' and ``\myvalue{San} \myvalue{Francisco}'', respectively. Only after all slot values are filled, the system can call the appropriate API to actually perform the intended action (e.g., reserving a table at a restaurant).
Thus, the extraction of slot values from natural languages utterances (i.e., slot filling) is a critical step to the success of a dialog system.

\begin{figure*}[t!]
  \centering
  \includegraphics[width=0.92\linewidth]{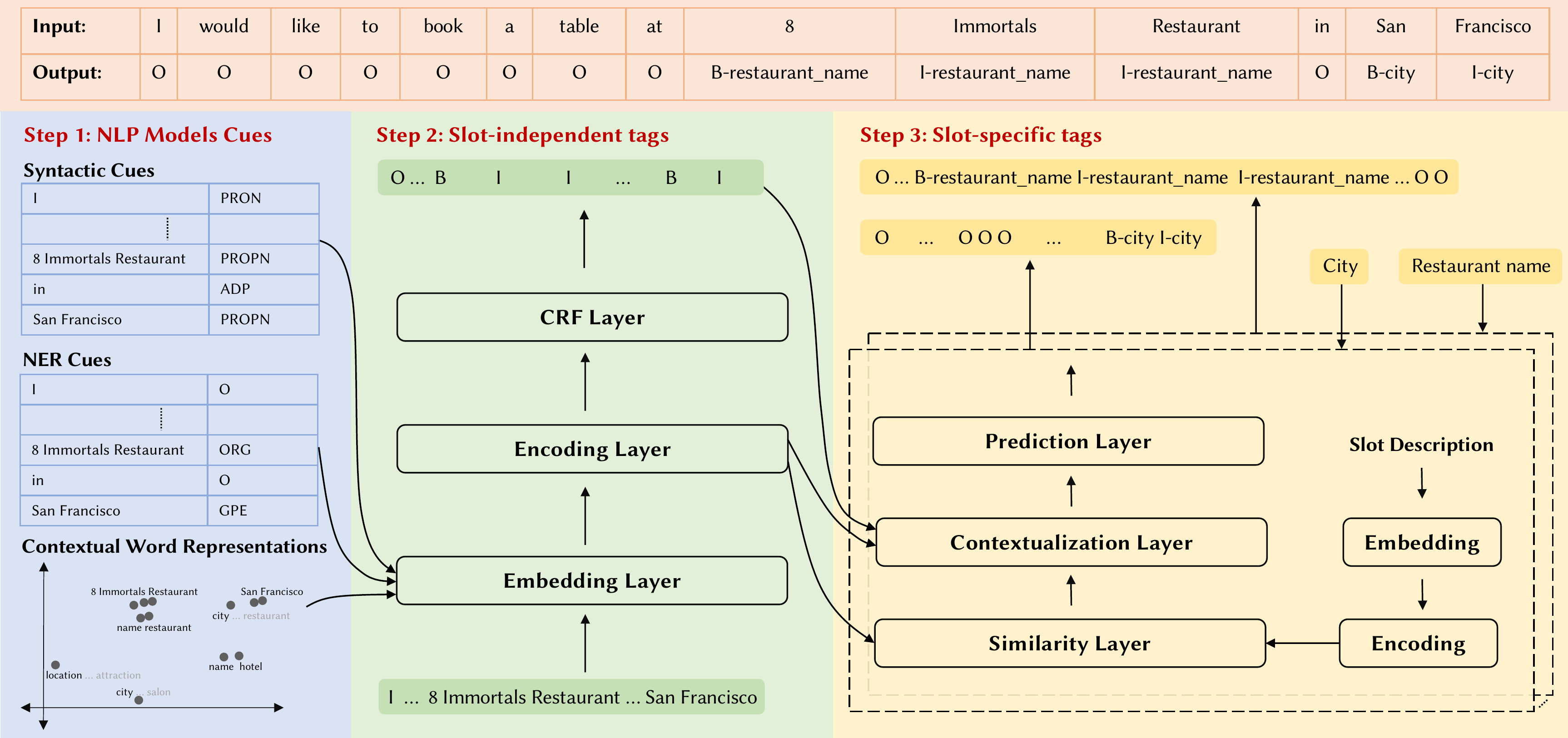}
  \caption{Overview of LEONA with an example utterance and its words' label sequence (following the IOB scheme). 
  }
  \vspace{-10pt}
  \label{fig:intro}
\end{figure*}

Slot filling is an important and challenging task that tags each word subsequence in an input utterance with a slot label (see Figure~\ref{fig:intro} for an example). Despite the challenges, 
supervised approaches have shown promising results for the slot filling task~\cite{goo2018slot,zhang2018joint,young2002talking,bellegarda2014spoken,mesnil2014using,kurata2016leveraging,hakkani2016multi,xu2013convolutional}.
The disadvantage of supervised methods is the unsustainable requirement of having massive labeled training data for each domain; the acquisition of such data is laborious and expensive.
Moreover, in practical settings, new unseen domains (with unseen slot types) emerge only after the deployment of the dialog system, rendering supervised models ineffective.
Consequently, models with capabilities to seamlessly adapt to new unseen domains are indispensable to the success of dialog systems. Note that unseen slot types do not have any training data, and the values of seen slots may be present in different contexts in new domains (rendering their training data from other seen domains irrelevant).
Filling slots in settings where new domains emerge after deployment is referred to as zero-shot slot filling~\cite{bapna2017towards}.
\myspecial{Alexa} \myspecial{Skills} and \myspecial{Google} \myspecial{Actions}, where developers can integrate their novel content and services into a virtual assistant are a prominent examples of scenarios where zero-shot slot filling is crucial.

There has been little research on zero-shot slot filling, and existing works presented limited experimental evaluation results. To the best of our knowledge, existing models were evaluated using a single public dataset.
Recently, the authors in~\cite{shah2019robust} proposed a cross-domain zero-shot adaptation for slot filling by utilizing example slot values. 
Due to the inherent variance of slot values, this framework faces difficulties in capturing the full slot value (e.g., ``\myvalue{8} \myvalue{Immortals} \myvalue{Restaurant}'' for slot type ``\myvalue{restaurant\_name}'' in Figure~\ref{fig:intro}) in unseen domains.
Coach~\cite{liu2020coach} proposed to address the issues in ~\cite{shah2019robust,bapna2017towards} with a coarse-to-fine approach. 
Coach~\cite{liu2020coach}  uses the seen domain data to learn templates for the slots based on whether the words are slot values or not.
Then, it determines a slot type for each identified slot value by matching it with the representation of each slot type description.
The diversity of slot types across different domains makes it practically impossible for Coach to learn general templates that are applicable to all new unseen domains; for example, ``\myvalue{book}'' and ``\myvalue{table}'' can be slot values in an e-commerce domain, but not in the restaurant booking domain.

We propose an end-to-end model \ourmodel\footnote{\textbf{L}inguistically-\textbf{E}nriched and c\textbf{ON}text-\textbf{A}ware}$^,$\footnote{Source code coming soon} that 
relies on the power of domain-independent linguistic features and contextual representations from pre-trained language models (LM), and context-aware utterance-slot similarity features.
{\ourmodel} works in three steps as illustrated in Figure~\ref{fig:intro}.
Step one leverages pre-trained Natural Language Processing (NLP) models that provide additional domain-oblivious and context-aware information to initialize our embedding layer.
Specifically, Step one uses \myNum{i}~syntactic cues through part of speech (POS) tags that provide information on the possibility of a word subsequence being a slot value (e.g., proper nouns are usually slot values); \myNum{ii}~off-the-shelf Named Entity Recognition (NER) models that provide complementary and more informative tags (e.g., geo-political entity tag for ``\myvalue{San} \myvalue{Francisco}''); and \myNum{iii}~a deep bidirectional pre-trained LM (ELMo)~\cite{Peters:2018} to generate contextual character-based word representations that can handle unknown words that were never seen during training. 
Moreover, the pre-trained ELMo~\cite{Peters:2018} with appropriate fine-tuning has provided state-of-the-art (SOTA) results on many NLP benchmarks~\cite{rajpurkar2016squad,bowman2015large,he2017deep,pradhan2012conll,sang2003introduction,socher2013recursive}.
Combined, these domain-independent sources of rich semantic information provide a robust initialization for the embedding layer to better accommodate unseen words (i.e., never seen during training), which greatly facilitates zero-shot slot filling.

Step two fine-tunes the semantically rich information from Step one by accounting for the temporal interactions among the utterance words using bi-directional Long Short Term Memory network~\cite{hochreiter1997long} that effectively transfers rich semantic information from NLP models.
This step produces slot-independent tags (i.e., Inside Outside Beginning \myspecial{IOB}), which provide complementary cues at the word subsequence level (i.e., hints on which word subsequences constitute slot values) using a Conditional Random Field (CRF)~\cite{lafferty2001conditional}.
Step three, which is the most critical step, learns a generalizable context-aware similarity function between the utterance words and those of slot descriptions from seen domains, and exploits the learned function in new unseen domains to highlight the features of the utterance words that are contextually relevant to a given slot.
This step also jointly contextualizes the multi-granular information produced at all steps.
Finally, CRF is employed to produce slot-specific predictions for the given utterance words and slot type.
This step is repeated for every relevant slot type, and the predictions are combined to get the final sequence labels. In our example in Figure~\ref{fig:intro}, the predictions for ``\myvalue{restaurant\_name}'' and ``\myvalue{city}'' are combined to produce the final sequence labels shown in the figure.

In summary, this work makes the following contributions:
\begin{itemize}[leftmargin=1.2\parindent,labelindent=-1pt, itemsep=-1pt]

    \item We propose an end-to-end model for zero-shot slot filling that effectively captures context-aware similarity between utterance words and slot types, and integrates contextual information across different levels of granularity, leading to outstanding zero-shot capabilities. 
    
    \item We demonstrate that pre-trained NLP models can provide additional domain-oblivious semantic information, especially for unseen concepts. 
    To the best of our knowledge, this is the first work that leverages the power of pre-trained NLP models for zero-shot slot filling.
    This finding might have positive implications for other zero-shot NLP tasks.
    
    \item We conduct extensive experimental analysis using four public datasets: \myspecial{SNIPS}~\cite{coucke2018snips}, \myspecial{ATIS}~\cite{liu2019benchmarking}, \myspecial{MultiWOZ}~\cite{zang-etal-2020-multiwoz} and \myspecial{SGD}~\cite{rastogi2019towards}, and show that our proposed model consistently outperforms SOTA models in a wide range of experimental evaluations on unseen domains. To the best of our knowledge, this is first work that comprehensively evaluates zero-shot slot filling models on many datasets with diverse domains and characteristics.
\end{itemize}

\section{Preliminaries}
\label{problem}
\subsection{Problem Formulation}

Given an utterance with $\bigJ$ words $\bigX_\smalli = (\smallx_\one, \smallx_\two,\cdots, \smallx_\bigJ )$, a slot value is a span of words $(\smallx_\smalle,\cdots,\smallx_\smallf)$ such that $0 \leq \smalle \leq \smallf \leq \bigJ$, that is associated with a slot type. 
Slot filling is a sequence labeling task that assigns the labels $\bigY_\smalli = (\smally_\one, \smally_\two,\cdots, \smally_\bigJ )$ to the input $\bigX_\smalli$, following the \myspecial{IOB} labeling scheme~\cite{ramshaw-marcus-1995-text}. 
Specifically, the first word of a slot value associated with slot type $\bigS_\smallr$ is labeled as \myvalue{B}-$\bigS_\smallr$, the other words inside the slot value are labeled as \myvalue{I}-$\bigS_\smallr$, and non-slot words are labeled as \myvalue{O}.
Let
$\bigD_{\smallc} = \{\bigS_\one, \bigS_\two, \dots\}$,
be the set of slot types in domain $\smallc$.
Let $\bigD_{\subvalue{SEEN}} = \{\bigD_\one,\cdots,\bigD_\smalll\}$ be a set of seen domains and
$\bigD_{\subvalue{UNSEEN}} = \{\bigD_{\smalll+\one}, \cdots,\bigD_\smallz\}$  be a set of unseen domains where $\bigD_{\subvalue{SEEN}}\cap \bigD_{\subvalue{UNSEEN}} = \varnothing$.
Let $\{(\bigX_\smalli, \bigY_\smalli)\}_{\smalli=1}^\smalln$ be a set of training utterances labeled at the word level such that the slot types in $\bigY_\smalli$ are in $\bigD_\smallp \in \bigD_{\subvalue{SEEN}}$.
In traditional (i.e., supervised) slot filling, the domains of test utterances belong to $\bigD_{\subvalue{SEEN}}$, whereas in zero-shot slot-filling, the domains of test utterances belong to $\bigD_{\subvalue{UNSEEN}}$;
an utterance belongs to a domain if it contains slot values that correspond to slot types from this domain.
Note that in zero-shot slot filling, the output slot types belong to either seen or unseen domains (i.e., in $\bigD_\smallp \in \bigD_{\subvalue{SEEN}} \cup \bigD_{\subvalue{UNSEEN}}$).
We focus on zero-shot slot filling in this work.

\subsection{Pre-trained NLP Models}
In this work, we utilize several pre-trained NLP models that are readily available. Specifically, we use: Pre-trained POS tagger, Pre-trained NER model, and Pre-trained ELMo.
The cues provided by POS/NER tags and ELMo embeddings are supplementary in our model, and they are further fine-tuned and contextualized using the available training data from seen domains.
Next, we provide a brief overview of these models.

\stitle{Pre-trained POS tagger.}
This model labels an utterance with part of speech tags, such as \myspecial{PROPN}, \myspecial{VERB}, and \myspecial{ADJ}. 
POS tags provide useful syntactic cues for the task of zero-shot slot filling, especially for unseen domains.
{\ourmodel} learns general cues from the language syntax about how slot values are defined in one domain, and transfers this knowledge to new unseen domains because POS tags are domain and slot type independent.
For example, proper nouns are usually values for some slots. In this work, we employ SpaCy's pre-trained POS tagger\footnote{\url{https://spacy.io/api/annotation\#pos-tagging}}, that has shown production level accuracy. 

\stitle{Pre-trained NER model.}
This model labels an utterance with \myspecial{IOB} tags for four entity types: \myspecial{PER}, \myspecial{GPE}, \myspecial{ORG}, and \myspecial{MISC}. The NER model provides information at a different granularity, which is generic and domain-independent.
Although the NER model provides tags for a limited set of entities and the task of slot filling encounters many more entity types, we observe that many, but not all, slots can be mapped to basic entities supported by the NER model.
For instance, names of places or locations are referred to as ``\myspecial{GPE}'' (i.e., geo-political entity or location) by the NER model, whereas in the task of the slot filling, there may be a location of a hotel, restaurant, salon, or some place the user is planing to visit. 
It remains challenging to assign the name of the location to the correct corresponding entity/slot in the zero-shot fashion. Moreover, NER models can not identify all slots/entities that slot filling intends to extract, resulting in a low recall. Yet, cues from NER model are informative and helpful in reducing the complexity of the task.
In this work, we employ SpaCy's pre-trained NER model\footnote{\url{https://spacy.io/api/annotation\#named-entities}}.

\stitle{Pre-trained ELMo.}
Pre-trained language models (i.e., ELMo) are trained on huge amounts of text data in an unsupervised fashion. These models have billions of parameters and thereby capture general semantic and syntactic information in an effective manner. 
In this work, we employ the deep bidirectional language model ELMo to provide contextualized word representations that capture complex syntactic and semantic features of words based on the context of their usage, unlike fixed word embeddings (i.e., GloVe~\cite{pennington2014glove} or Word2vec~\cite{mikolov2013efficient}) which do not consider context.
Furthermore, these representations are purely character based and are robust for words unseen during training, which makes them suitable for the task of zero-shot slot filling.

\subsection{Conditional Random Fields}
Conditional Random Fields (CRFs)~\cite{sutton2006introduction} have been successfully applied to various sequence labeling problems in natural language processing such as POS tagging~\cite{cutting1992practical}, shallow parsing~\cite{sha2003shallow}, and named entity recognition~\cite{settles2004biomedical}.
To produce the best possible label sequence for a given input, CRFs incorporate the context and dependencies among predictions. In this work, we employ linear chain CRFs that are trained by estimating maximum conditional log-likelihood. In its simplest form, it estimates a transition cost matrix of size,  \myvalue{num\_tags} $\times$ \myvalue{num\_tags}, where the value at the indices [\myvalue{i},~\myvalue{j}] represents the likelihood of transitioning from the \myvalue{j}-th tag to the \myvalue{i}-th tag. Moreover, it allows enforcing constraints in a flexible way (e.g., tag ``\myvalue{I}'' can not be preceded by tag ``\myvalue{O}'').
\section{Approach}
\label{model}

\begin{figure*}[t!]
  \centering
  \includegraphics[width=0.85\linewidth]{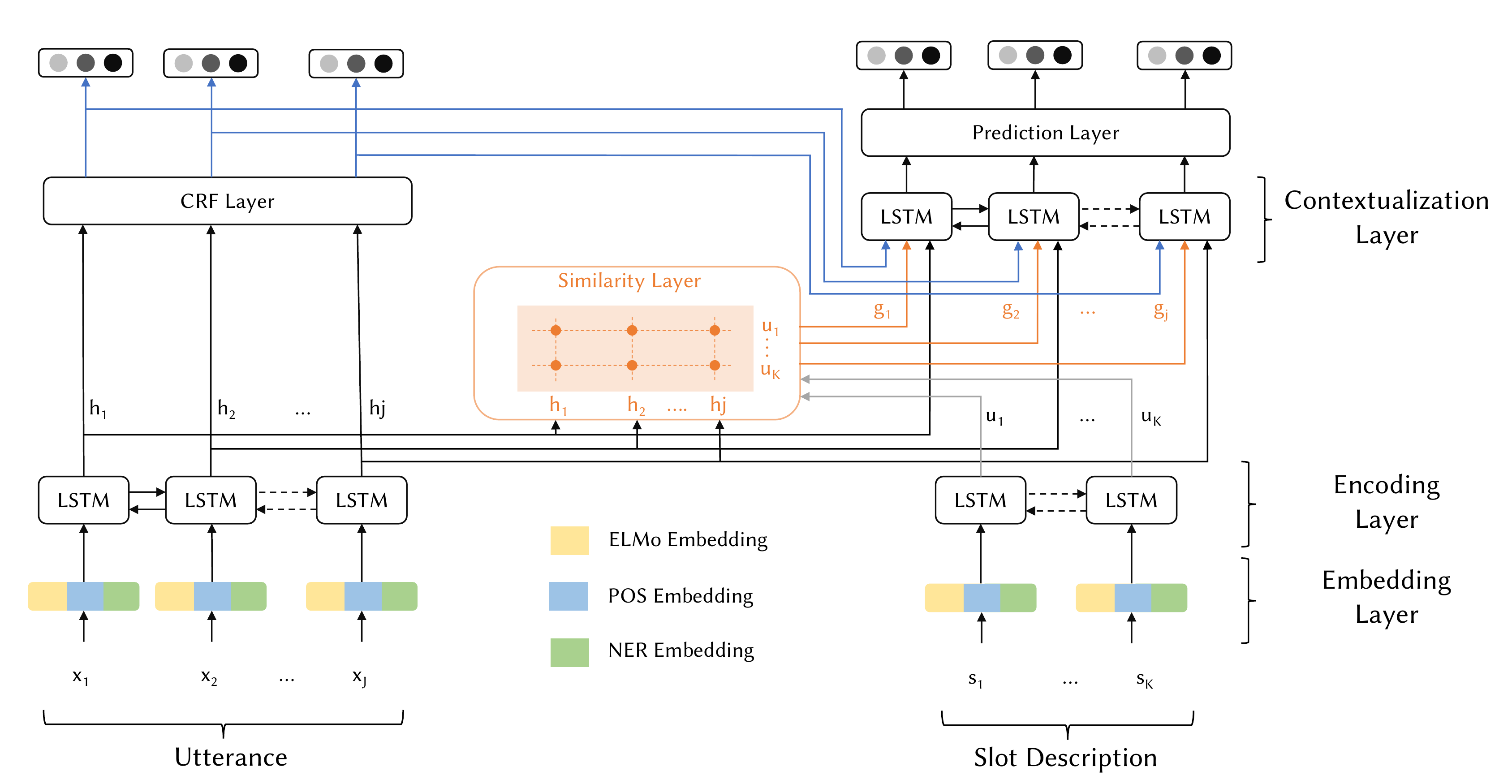}
  \caption{Illustration of the layers in our model LEONA.}
  \label{fig:model}
\end{figure*}

Our model {\ourmodel} is an end-to-end neural network with six layers that collectively realize the conceptual three steps in Figure~\ref{fig:intro}.
Specifically, the Embedding layer realizes \myspecial{Step one} and it also jointly realizes \myspecial{Step two} together with the Encoding and the CRF layers. The Similarity, Contextualization, and Predication layers realize \myspecial{Step three}.
We briefly summarize each layer below, and we describe each layer in detail in the subsequent subsections.
The Embedding layer maps each word to a vector space; this layer is responsible for embedding the words from both the utterance and the slot description. 
The Encoding layer uses bi-directional LSTM networks to refine the embeddings from the previous layer by considering information from neighboring words. This layer encodes utterances as well as slot descriptions.
The CRF layer uses utterance encodings and makes slot-independent predictions (i.e., \myspecial{IOB} tags) for each word in the utterance by considering dependencies between the predictions and taking context into account.
The Similarity layer uses utterance and slot description encodings to compute an attention matrix that captures the similarities between utterance words and a slot type, and signifies feature vectors of the utterance words relevant to the slot type.
The Contextualization layer uses representations from different granularities and contextualizes them for slot-specific predictions by employing bi-directional LSTM networks; specifically, it uses representations from the Similarity layer, the Encoding layer, and the \myspecial{IOB} predictions produced by the CRF layer. 
The Prediction layer employs another CRF to make slot-specific predictions (i.e., \myspecial{IOB} tags for a given slot type) based on the input from the contextualization layer.
Note that the prediction process is repeated for all the relevant slot types and its outputs are combined to produce the final label for each word.

\subsection{Embedding Layer}
This layer maps each word in the input utterance to a high-dimensional vector space. Three complementary embeddings are utilized: \myNum{i} word embedding of the POS tags for the input words, \myNum{ii} word embedding of the NER tags for the input word, and \myNum{iii} contextual word embedding from the pre-trained ELMo model.
Then, we employ a two-layer Highway Network~\cite{srivastava2015highway} to combine the three embeddings for each word in an effective way. Highway Networks have been shown to have better performance than simple concatenation. They produces a $\smalld$-dimensional vector for each word. Specifically, the embedding layer produces $\bigX \in \realR^{\smalld \times \bigJ}$ for the given utterance $\{\smallx_\one, \smallx_\two, \cdots, \smallx_\bigJ\}$ with $\bigJ$ words, and $\bigS \in \realR^{\smalld \times \bigK}$ for the given slot description $\{\smalls_\one, \smalls_\two, \cdots, \smalls_\bigK\}$ with $\bigK$ words. This representation gets fine-tuned and contextualized in the next layers.

\subsection{Encoding Layer}
We use a bi-directional LSTM network to capture the temporal interactions between input words. At time-step $\smalli$, we compute hidden states for the input utterance as follows:
\begin{equation*}
  \overrightarrow{\smallh}_\smalli = \text{LSTM} (\overrightarrow{\smallh}_{\smalli-\one}, \bigX_{:\smalli})
  \label{eq:encoder-right}
\end{equation*}
\begin{equation*}
  \overleftarrow{\smallh}_\smalli = \text{LSTM} (\overleftarrow{\smallh}_{\smalli-\one}, \bigX_{:\smalli})
  \label{eq:encoder-left}
\end{equation*}
Then, we concatenate the output of the hidden states $\overrightarrow{\smallh}_\smalli$ and $\overleftarrow{\smallh}_\smalli$ to get the bi-directional hidden state representation, $\smallh_\smalli = [\overrightarrow{\smallh}_\smalli ; \overleftarrow{\smallh}_\smalli] \in \realR^{\two\lstmd}$. This layer produces $\bigH \in \realR^{\two\lstmd \times \bigJ}$ from the context word vectors $\bigX$ (i.e., for utterance). Essentially, every column of the matrix represents the fine-tuned context-aware representation of the corresponding word.
A similar mechanism is employed to produce $\bigU \in \realR^{\two\lstmd \times \bigK}$ from word vector $\bigS$ (i.e., for slot description).

\subsection{CRF Layer}
The task of the CRF layer is to predict one of three slot-independent tags (i.e., \myspecial{I}, \myspecial{O}, or \myspecial{B}) for each word based on utterance contextual representation $\bigH = \{ \smallh_\one, \smallh_\two, \cdots, \smallh_\bigJ \}$ produced by the encoding layer. Let $\bigY$ refer to a sequence label, and the set of all possible state sequences is $\bigC$. For the given input sequence $\bigH$, the conditional probability function for the CRF, $P(\bigY|\bigH; W, b)$, over all possible label sequences $\bigY$ is computed as follows:
\begin{equation*}
P(\bigY|\bigH; W, b) = \frac{\prod\limits_{\smalli=\one}^\bigJ \theta_\smalli (\smally_{\smalli-\one}, \smally_\smalli, \bigH)}{\sum \limits_{\smally' \in \bigC} \prod\limits_{\smalli=\one}^\bigJ \theta_\smalli (\smally_{\smalli-\one}', \smally_\smalli', \bigH)}  
\label{eq:crf}
\end{equation*}
where $\theta_\smalli (\smally_{\smalli-\one}', \smally_\smalli', \bigH) = \exp(W_{\smally',\smally}^T \smallh_\smalli + b_{\smally',\smally})$ is a trainable function, that has $W_{y',y}^T$ weight and $b_{y',y}$ bias matrices for the label pair $(y', y)$. 

Note that the slot-independent predictions also represent the output of \myspecial{Step} \myspecial{two}; i.e., information about utterance words at a different granularity than the initial cues from NLP models. 
Essentially, \myspecial{Step} \myspecial{two} learns general patterns of slot values from seen domains irrespective of slot types, and transfers this knowledge to new unseen domains and their slot types. Since it is hard to learn general templates of slot values that are applicable to all unseen domains, we do not use these slot-independent predictions to predict slot-specific tags.
Instead, we pass this information to the contextualization layer for further fine-tuning.

\subsection{Similarity Layer}
The similarity layer highlights the features of each utterance word that are important for a given slot type by employing attention mechanisms. The popular attention methods~\cite{weston2014memory,bahdanau2014neural,liu2016attention} that summarize the whole sequence into a fixed length feature vector are not suitable for the task at hand, i.e., per word labeling. 
Alternatively, we compute the attention vector at each time step, i.e., attention vector for each word in the utterance. The utterance encoding $\bigH \in \realR^{\two\lstmd \times \bigJ}$ and slot description encoding $\bigU \in \realR^{\two\lstmd \times \bigK}$ metrics are input to this layer, that are used to compute a similarity matrix $\bigA \in \realR^{\bigJ \times \bigK}$ between the utterance and slot description encodings. $\bigA_{\smallj\smallk}$ represents the similarity between $\smallj$-th utterance word and $\smallk$-th slot description word. We compute the similarity matrix, as follows:
\begin{equation*}
\bigA_{\smallj\smallk} = \alpha (\bigH_{:\smallj}, \bigU_{:\smallk}) \in \realR
\label{eq:similarity}
\end{equation*}
where $\alpha$ is a trainable function that captures the similarity between input vectors $\bigH_{:\smallj}$ and $\bigU_{:\smallk}$, where $\bigH_{:\smallj}$ and $\bigU_{:\smallk}$ are $\smallj$-th and $\smallk$-th column-vectors of $\bigH$ and $\bigU$, respectively.
$\alpha (\smallh, \smallu) = w^\top_{(\smalla)} [ \smallh \oplus \smallu \oplus \smallh \otimes \smallu]$, where $\oplus$ is vector concatenation, $\otimes$ is element-wise multiplication, and $w_{(\smalla)}$ is a trainable weight vector.

The similarity matrix $\bigA$ is used to capture bi-directional interactions between the utterance words and the slot type.
First we compute attention that highlights the words in the slot description that are closely related to the utterance. At time-step $\smallt$, we compute it as follows: 
$
\bigU'_{:\smallt} = \sum _\smallk
\smallv_{\smallt\smallk} \bigU_{:\smallk}
$
where $\smallv_\smallt = \text{softmax}(\bigA_{\smallt:}) \in \realR^\bigK$ is the attention weight for slot description computed at time-step $\smallt$ and $\sum \smallv_{\smallt\smallk} = 1$ for all $\smallt$. $\bigU' \in \realR^{\two\lstmd\times\bigJ}$ represents the attention weights for the slot description with respect to all the words in the utterance. 
Basically, every column of the matrix represents closeness of the slot description with the corresponding utterance word.
Then, attention weights that signify the words in the utterance that have the highest similarity with the slot description are computed as follows:
$
\smallh' = \sum _\smallj
\smallb_{\smallj} \bigH_{:\smallj}
$
where $\smallb = \text{softmax}(\text{max}_\text{col}(\bigA)) \in \realR^\bigJ$ and \text{max} is operated across columns, and $\bigH' \in \realR ^ {\two\lstmd\times\bigJ}$ is obtained by tiling $\smallh'$ across columns.

We highlight that $\bigU'$ represents features that highlight important slot description words with closely similar words of utterance, and $\bigH'$ highlights features of the utterance with high similarity with the slot description, computed based on the similarity matrix $\bigA$, that itself has been computed based the contextual representations of the utterance ($\bigH$) and slot description ($\bigU$) generated by the encoding layer that considers surrounding words (i.e., employing bi-LSTM) to generate the representations. Finally, $\bigU'$ and $\bigH'$ are concatenated to produce $\bigG \in \realR ^{\four\lstmd\times\bigJ}$, where every column of the matrix represents rich bi-directional similarity features of the corresponding utterance word with the slot description.

Essentially, this layer learns a general context-aware similarity function between utterance words and a slot description from seen domains, and it exploits the learned function for unseen domains. Due to the general nature of the similarity function, this layer also facilitates the identification of slot values in cases when \myspecial{Step} \myspecial{two} fails to correctly identify domain-independent slot values.

\subsection{Contextualization Layer}
This layer is responsible for contextualizing information from different granularities. Specifically, the utterance encodings from the Encoding layer, the bi-directional similarity between the utterance and the slot description from  the Similarity layer, and the slot-independent \myspecial{IOB} predictions from the CRF layer are passed as input. This layer employs $\two$ stacked layers of bi-directional LSTM networks to contextualize all the information by considering the relationships among neighbouring words' representations. It generates high quality features for the prediction layer; specifically, the features are $\in \realR^{\two\lstmd\times\bigJ}$, where each column represents the $\two\lstmd$-dimensional features for the given word in the utterance.

\subsection{Prediction Layer}
The contextualized features are passed as input to this layer, and it is responsible for generating slot-specific predictions for the given utterance and slot type. First, it passes these features through $\two$ linear layers with \text{ReLU} activation. Then a CRF is employed to make structured predictions, as briefly explained in the CRF layer.
The prediction process is done for each of the relevant slot types (i.e., slot types in the respective domain) and the resulting label sequences are combined to produces the final label for each word. 
Note that if the model made two or more conflicting slot predictions for a given sequence of words, we pick the slot type with the highest prediction probability.

\subsection{Training the Model}
The model has two trainable components: the slot-independent \myspecial{IOB} predictor and the slot-specific \myspecial{IOB} predictor. We jointly train both components by minimizing the negative log likelihood loss of both components over our training examples.
The training data is prepared as follows. The training examples are of the form $(\bigX_\smalli, \bigS_\smallr, \bigY'_\smalli, \bigY''_{\smalli\smallr})$, where $\bigX_\smalli$ represents an utterance, $\bigS_\smallr$ represents a slot type, $\bigY'_\smalli$ represents slot-independent \myspecial{IOB} tags for the given utterance $\bigX_\smalli$, and $\bigY''_{\smalli\smallr}$ represents slot-specific \myspecial{IOB} tags for the given utterance $\bigX_\smalli$ and slot type $\bigS_\smallr$. 
For a sample from the given dataset of the form $(\bigX_\smalli, \bigY_\smalli)$ that has values for $\smallm$ slot types, first slot-indepedent \myspecial{IOB} tags $\bigY'_\smalli$ are generated by removing slot type information. Then, we generated $\smallm$ positive training examples by setting each of $\smallm$ slot types as $\bigS_\smallr$ and generating corresponding label $\bigY''_{\smalli\smallr}$ (i.e., slot-specific tags for slot type $\bigS_\smallr$). Finally, $\smallq$ negative examples are generated, where such slot types are chosen which are not present in the utterance. For example, the utterance in Figure~\ref{fig:intro} ``I would like to book a table at 8 Immortals Restaurant in San Francisco'' has true labels as ``O O O O O O O O B-restaurant\_name I-restaurant\_name I-restaurant\_name O B-city I-city''. The positive training examples would be: ($\cdots$, ``\myvalue{restaurant\_name}'', ``O O O O O O O O B I I O B I'', ``O O O O O O O O \myvalue{B} \myvalue{I} \myvalue{I} O O O'') and ($\cdots$, ``\myvalue{city}'', $\cdots$, ``O O O O O O O O O O O O \myvalue{B} \myvalue{I}'').
Whereas the negative examples can be as follows: ($\cdots$, ``\myvalue{salon\_name}'', $\cdots$, ``O O O O O O O O O O O O O O''), ($\cdots$, ``\myvalue{cuisine}'', $\cdots$, $\cdots$), ($\cdots$, ``\myvalue{phone\_number}'', $\cdots$, $\cdots$), and so on.
Note that slot types are shown in the above example for brevity, the slot descriptions are used in practice.
\section{Experimental Setup}
\label{experiments}

\begin{table}[t!]
\centering
\caption{Dataset statistics.}
\label{tab:dataset}
\begin{tabular}{lcccc}
\toprule
\textbf{Dataset}         & \textbf{SNIPS} & \textbf{ATIS} & \textbf{MultiWOZ}  & \textbf{SGD}
\\ \hline
Dataset Size    & $14.5$K     & $5.9$K & $67.4$K    &$188$K   \\
Vocab. Size & $12.1$K     & $1$K & $10.5$K   &$33.6$K  \\
Avg. Length &  $9.0$     & $11.1$ & $13.3$   &$13.8$  \\
\# of Domains    & $6$     & $1$ & $8$   &$20$  \\
\# of Intents    & $7$     & $18$ & $11$   &$46$  \\
\# of Slots    & $39$     & $83$ & $61$   &$240$  \\
\bottomrule
\end{tabular}
\end{table}

In this section, we describe the datasets, evaluation methodology, competing methods, and the implementation details of our model {\ourmodel}.

\subsection{Datasets}
\label{datasets}
We used four public datasets to evaluate the performance of our model \ourmodel: SNIPS Natural Language Understanding benchmark (\myspecial{SNIPS})~\cite{coucke2018snips}, Airline Travel Information System (\myspecial{ATIS})~\cite{liu2019benchmarking}, Multi-Domain Wizard-of-Oz (\myspecial{MultiWOZ})~\cite{zang-etal-2020-multiwoz}, and Dialog System Technology Challenge 8, Schema Guided Dialogue (\myspecial{SGD})~\cite{rastogi2019towards}. To the best of our knowledge, this is first work to comprehensively evaluate zero-shot slot filling models on a wide range of public datasets. Table~\ref{tab:dataset} presents important statistics about the datasets.

\stitle{SNIPS.}
A crowd-sourced single-turn Natural Language Understanding (NLU) benchmark widely used for slot filling. It has $39$ slot types across $7$ intents from different domains. Since this dataset does not have slot descriptions, we used tokenized slot names as the descriptions (e.g., for slot type ``\myspecial{playlist\_owner}'', we used ``\myspecial{playlist} \myspecial{owner}'' as its description).

\stitle{ATIS.}
A single-turn dataset that has been widely used in slot filling evaluations.
It covers $83$ slot types across $18$ intents from a single domain. Many of the intents do not have many utterances, so all the intents having less than $100$ utterances are combined into a single intent ``\myspecial{Others}'' in our experiments. Moreover, similarly to \myspecial{SNIPS} dataset, we used the tokenized versions of the slot names as slot descriptions.

\stitle{MultiWOZ.}
A well-known dataset that has been widely used for the task of dialogue state tracking. In this work, we used the most recent version of the dataset (i.e., \myspecial{MultiWOZ}$2.2$). In its original form, it contains dialogues between users and system. For the task of slot filling, we take all the user utterances and system messages that mention any slot(s), and shuffle the order to make it as if it were a single-turn dataset to maintain consistency with the previous works.
For experiments in this work, utterances with intents that have less than $650$ ($< 1\%$ of the dataset) utterances are grouped into the intent ``\myspecial{Others}''.

\stitle{SGD.}
A recently published comprehensive dataset for the eighth Dialog System Technology Challenge; it contains dialogues from $20$ domains with a total of $46$ intents and $240$ slots. \myspecial{SGD} was originally proposed for dialogue state tracking. This dataset is also pre-processed to have single-turn utterances labeled for slot filling.
Moreover, we merge utterances from domains that have no more than $1850$ ($< 1\%$ of the dataset) utterances, and we name the resulting domain ``\myspecial{Others}''.

Since not all datasets provide a large enough number of domains, we do the splits in our experiments based on intents instead of domains for datasets that have more intents than domains. That is, we consider intents as domains for \myspecial{SNIPS}, \myspecial{ATIS}, and \myspecial{MultiWOZ}.

\subsection{Evaluation Methodology}
\label{testing}

We compute the slot F1 scores\footnote{Standard CoNLL evaluation script is used to compute slot F1 score.} and present evaluation results for the following settings:

\stitle{Train on all except target intent/domain.}
This is the most common setting that previous works~\cite{liu2020coach,shah2019robust,bapna2017towards} have used for evaluation. A model is trained on all intents/domains except a single target intent/domain. For example, for \myspecial{SNIPS} dataset the model is trained on all intents except a target intent ``\myspecial{AddToPlatlist}'' that is used for testing the model's capabilities in the zero-shot fashion.
This setup is repeated for every single intent in the dataset.
The utterances at test time only come from a single intent/domain (or ``\myspecial{Others}'') which makes this setting less challenging.

\stitle{Train on a certain percentage of intents/domains and test on the rest.} 
This is a slightly more challenging setting where test (i.e., unseen in training) intent/domains are usually from multiple unseen new intents/domains. We vary the number of training (i.e., seen) and testing (i.e., unseen) intents/domains to comprehensively evaluate all competing models. In this setting, we randomly select $\approx 25\%$, $\approx 50\%$, and $\approx 75\%$ of the intents/domains for training and the rest for testing, and report average results over five runs. 

\stitle{Train on one dataset and test on the rest of the datasets.}
This is the most challenging setting, where models are trained on one dataset and tested on the remaining datasets. For example, we train on the \myspecial{SGD} dataset and test on \myspecial{SNIPS}, \myspecial{ATIS}, and \myspecial{MultiWOZ} datasets. Similarly, we repeat the process for every dataset.
Since datasets are very diverse (i.e., in terms of domains, slot types and user's expressions), this setting can be thought of as a ``\myspecial{in} \myspecial{the} \myspecial{wild}''~\cite{dhall2017individual} setting, which resembles real-world zero-shot slot filling scenarios to a large degree.

\subsection{Competing Methods}
\label{baselines}
We compare against the following state-of-the-art (SOTA) models:
\begin{description}[leftmargin=1.2\parindent,labelindent=-3.5pt, itemsep=-1pt]
\item \textbf{Coach~\cite{liu2020coach}.}
This model proposes to handle the zero-shot slot filling task with a coarse-to-fine procedure. It first identifies the words that constitute slot values. Then, based on the identified slot values, it tries to assign these values to slot types by matching the identified slot values with the representation of each slot description. We use their best model, i.e., Coach+TR, that employs template regularization but we call it Coach for simplicity.

\item \textbf{RZS~\cite{shah2019robust}.}
This work proposes a zero-shot adaption for slot filling by utilizing example values of each slot type. It employs character and word embedding of the utterance and slot descriptions, which are then concatenated with the averaged slot example embeddings, and passed through a bidirectional LSTM network to get the final prediction for each word in the utterance.

\item \textbf{CT~\cite{bapna2017towards}.}
This model fills slots for each slot type individually. Character and word-level representations are concatenated with the slot type representation (i.e., embeddings) and an LSTM network is used to make the predictions for each word in the utterance for the given slot type.

\end{description}

Note that we do not compare against simple baselines such as BiLSTM-CRF~\cite{lample2016neural}, LSTM-BoE, and CRF-BoE~\cite{jha2018bag}  because they have been outperformed by the previous works we compare against.

\subsection{Implementation Details}
\label{implementation}
Our model uses $300$ dimensional embedding for POS and NER tags, and pre-trained ELMo embedding with $1024$ dimensions. The encoding and contextualization layers have two stacked layers of bi-directional LSTMs with hidden states of size $300$. The prediction layer has two linear layers with \text{ReLU} activation, and the CRF uses the ``\myspecial{IOB}'' labeling scheme. The model is trained with a batch size of $32$ for up to $200$ epochs with early stopping using Adam optimizer and a negative log likelihood loss with a scheduled learning rate, starting at $0.001$, and the model uses dropout rate of $0.3$ at every layer to avoid over-fitting. Whereas $\smallq$ is set to three for negative sampling.

\section{Results}
\label{results}

\begin{table}[t!]
\footnotesize
\caption{SNIPS dataset: Slot F1 scores for all competing models for target intents that are unseen in training.}
\label{tab:zs_snips}
\begin{tabular}{l|ccc|cc}
\toprule
\textbf{Target Intent $\downarrow$}  & \textbf{CT}     & \textbf{RZS}    & \textbf{Coach}    & \textbf{LEONA w/o IOB} & \textbf{LEONA}\\ \hline
AddToPlaylist    & 0.3882 & 0.4277 & 0.5090     & 0.5104 & \textbf{0.5115}     \\
BookRestaurant        & 0.2754 & 0.3068 & 0.3401    & 0.3405  & \textbf{0.4781}    \\
GetWeather           & 0.4645 & 0.5028 & 0.5047    & 0.5531 & \textbf{0.6677}     \\
PlayMusic            & 0.3286 & 0.3312 & 0.3201    & 0.3435  & \textbf{0.4323}    \\
RateBook             & 0.1454 & 0.1643 & 0.2206   & 0.2224  & \textbf{0.2318}     \\
SearchCreativeWork   & 0.3979 & 0.4445 & 0.4665    & 0.4671  & \textbf{0.4673}    \\
SearchScreeningEvent & 0.1383 & 0.1225 & 0.2563    & 0.2690 & \textbf{0.2872}      \\ \hline
Average              & 0.3055 & 0.3285 & 0.3739    & 0.3866 & \textbf{0.4394}   \\
\bottomrule
\end{tabular}
\end{table}

\begin{table}[t!]
\footnotesize
\caption{ATIS dataset: Slot F1 scores for all competing models for target intents that are unseen in training.}
\label{tab:zs_atis}
\begin{tabular}{l|ccc|cc}
\toprule
\textbf{Target Intent $\downarrow$} & \textbf{CT}     & \textbf{RZS}    & \textbf{Coach}    & \textbf{LEONA w/o IOB} & \textbf{LEONA}\\ \hline
Abbreviation          & 0.4163 & 0.5252 & 0.4804   & 0.4965   & \textbf{0.6405}    \\
Airfare               & 0.6549 & 0.5410  & 0.6929    & 0.7490 & \textbf{0.9492}      \\
Airline               & 0.7126 & 0.6354 & 0.7212    & 0.7762  & \textbf{0.8586}    \\
Flight                & 0.6530  & 0.7165 & 0.8072     & 0.8521 & \textbf{0.9070}     \\
Ground Service       & 0.4924 & 0.6452 & 0.7641     & 0.8463 & \textbf{0.8490}     \\
Others               & 0.4835 & 0.5169 & 0.6586    & 0.7749  & \textbf{0.8337}    \\ \hline
Average              & 0.5688 & 0.5967 & 0.6874   & 0.7492   & \textbf{0.8397}   \\
\bottomrule
\end{tabular}
\end{table}

We present in the next subsections quantitative and qualitative analysis of all competing models. We first present the quantitative analysis in Subsection~\ref{results-quantitative} and show that our model consistently outperforms the competing models in all settings. Furthermore, this subsection also has an ablation study that quantifies the role of each conceptual step in our model. We dig deeper into limitations of each competing model in our qualitative analysis in Subsection~\ref{results-qualitative}.

\subsection{Quantitative Analysis}
\label{results-quantitative}
\stitle{Train on all except target intent/domain.}
Tables~\ref{tab:zs_snips} ~\ref{tab:zs_atis}, ~\ref{tab:zs_multiwoz}, and ~\ref{tab:zs_sgd} present F1 scores for \myspecial{SNIPS}, \myspecial{ATIS}, \myspecial{MultiWOZ}, and \myspecial{SGD} datasets, respectively.
All models are trained on all the intents/domains except the target one that is used for zero-shot testing.
Our proposed approach is consistently better than SOTA methods. Specifically, it outperforms SOTA models by $17.52\%$, $22.15\%$, $17.42\%$, and $17.95\%$ on average for unseen intents/domains on \myspecial{SNIPS}, \myspecial{ATIS}, \myspecial{MultiWOZ}, and \myspecial{SGD} datasets, respectively.
We also present a variant of our model that does not employ ``\myspecial{IOB}'' tags from \myspecial{Step} \myspecial{two}, we call it {\ourmodel~\noiob}. Even this variant of our model outperforms all other SOTA models.
This performance gain over SOTA methods can be attributed to the pre-trained NLP models that provide meaningful cues for the unseen domains, the similarity layer that can capture the closeness of the utterance words with the given slot irrespective of whether it is seen or unseen, and the contextualization layer that uses all the available information to generate a rich context-aware representation for each word in the utterance.

\begin{table}[t!]
\footnotesize
\caption{MultiWOZ dataset: Slot F1 scores for all competing models for target intents that are unseen in training.}
\label{tab:zs_multiwoz}
\begin{tabular}{l|ccc|cc}
\toprule
\textbf{Target Intent $\downarrow$}    & \textbf{CT}     & \textbf{RZS}    & \textbf{Coach}    & \textbf{LEONA w/o IOB} & \textbf{LEONA} \\ \hline
Book Hotel           & 0.4577 & 0.3739 & 0.5866  & 0.6181  & \textbf{0.6446}    \\
Book Restaurant      & 0.3260  & 0.4200   & 0.4576  & 0.6268  & \textbf{0.6269}    \\
Book Train           & 0.4777 & 0.5269 & 0.6112  & 0.6317 & \textbf{0.7025}     \\
Find Attraction    &  0.2914 & 0.3489 & 0.3029  & 0.3787  & \textbf{0.3834}    \\
Find Hotel           & 0.4933 & 0.5920  & 0.7235  & 0.7673  & \textbf{0.8222}    \\
Find Restaurant      & 0.6420  & 0.6921 & 0.7671  & 0.7969  & \textbf{0.8338}    \\
Find Taxi            & 0.1459 & 0.1587 & 0.1260   & 0.1682   & \textbf{0.1824}   \\
Find Train           & 0.6344 & 0.4406 & 0.7754  & 0.8779 & \textbf{0.8811}     \\
Others               & 0.1205 & 0.0878 & 0.1201  & 0.1687  & \textbf{0.1721}    \\ \hline
Average              & 0.3988 & 0.4045 & 0.4967  & 0.5594 & \textbf{0.5832}  \\
\bottomrule  
\end{tabular}
\end{table}

\begin{table}[t!]
\footnotesize
\caption{SGD dataset: Slot F1 scores for all competing models for target domains that are unseen in training.}
\label{tab:zs_sgd}
\begin{tabular}{l|ccc|cc}
\toprule
\textbf{Target Domain $\downarrow$}    & \textbf{CT}     & \textbf{RZS}    & \textbf{Coach}    & \textbf{LEONA w/o IOB} & \textbf{LEONA}\\ \hline
Buses                & 0.4954 & 0.5443 & 0.6280   & 0.6364 & \textbf{0.6978}     \\
Calendar             & 0.5056 & 0.4908 & 0.6023  & 0.6216  & \textbf{0.7436}    \\
Events               & 0.5181 & 0.6324 & 0.5486  & 0.7405 & \textbf{0.7619}     \\
Flights              & 0.4898 & 0.4662 & 0.4898  & 0.4907   & \textbf{0.5901}   \\
Homes                & 0.4542 & 0.7159 & 0.6235  & 0.6927  & \textbf{0.7698}    \\
Hotels               & 0.4069 & 0.5681 & 0.7216  & 0.7266  & \textbf{0.7677}    \\
Movies               & 0.5100   & 0.3424 & 0.5537  & 0.5687 & \textbf{0.7285}     \\
Music                & 0.4111 & 0.6090  & 0.5786  & 0.7466 & \textbf{0.7613}     \\
RentalCars           & 0.4138 & 0.3399 & 0.6576  & 0.7344 & \textbf{0.7389}     \\
Restaurants          & 0.4620  & 0.3787 & 0.7195 & 0.7451   & \textbf{0.7574}    \\
RideSharing          & 0.6619 & 0.5312 & 0.7273 & 0.7656   & \textbf{0.8172}    \\
Services             & 0.6380  & 0.6381  & 0.7607   & 0.7628 & \textbf{0.8180}     \\
Travel               & 0.6556 & 0.6464 & 0.8403  & 0.9013 & \textbf{0.9234}     \\
Weather              & 0.4605 & 0.5180  & 0.6003  & 0.6178 & \textbf{0.8223}     \\
Others               & 0.4362 & 0.5312 & 0.4921  & 0.5129   & \textbf{0.5592}   \\ \hline
Average              & 0.5013 & 0.5302 & 0.6363  & 0.6842 & \textbf{0.7505}   \\
\bottomrule  
\end{tabular}
\end{table}

\begin{table*}[t!]
\caption{Averaged F1 scores for all competing models for seen and unseen slot types in the target unseen domains for SNIPS, ATIS, MultiWOZ, and SGD datasets.}
\label{tab:seen_unseen}
\begin{tabular}{l|cc|cc|cc|cc}
\toprule
\textbf{Method $\downarrow$}      & \multicolumn{2}{c|}{\textbf{SNIPS}} & \multicolumn{2}{c|}{\textbf{ATIS}} & \multicolumn{2}{c|}{\textbf{MultiWOZ}} & \multicolumn{2}{c}{\textbf{SGD}} \\ \hline
Slot Type $\rightarrow$   & Seen         & Unseen      & Seen        & Unseen      & Seen          & Unseen        & Seen        & Unseen     \\ \hline
CT          & 0.4407       & 0.2725      & 0.7552      & 0.4851      & 0.6062        & 0.3040        & 0.7362      & 0.3940     \\
RZS         & 0.4786       & 0.2801      & 0.8132      & 0.5143      & 0.6604        & 0.3301        & 0.7565      & 0.4478     \\
Coach    & 0.5173       & 0.3423      & 0.7742      & 0.7166      & 0.7034        & 0.4895        & 0.7996      & 0.6614     \\ \hline
\ourmodel~\noiob & 0.5292       & 0.3578      & 0.8155      & 0.7130      & 0.6651        & 0.5638        & 0.7986      & 0.7424     \\ 
\ourmodel        & \textbf{0.6354}       & \textbf{0.4006}      & \textbf{0.9588}      & \textbf{0.7524}      & \textbf{0.7765}        & \textbf{0.5962}        & \textbf{0.9192}      & \textbf{0.8167}     \\
\bottomrule 
\end{tabular}
\end{table*}

\begin{table*}[t!]
\caption{Averaged F1 scores for all competing models in the target unseen domains of all datasets. The train/test sets have variable number of intents/domains, which makes this setting more challenging.}
\label{tab:zs_percent}
\begin{tabular}{l|ccc|ccc|ccc|ccc}
\toprule
\textbf{Method $\downarrow$}      & \multicolumn{3}{c|}{\textbf{SNIPS}} & \multicolumn{3}{c|}{\textbf{ATIS}} & \multicolumn{3}{c|}{\textbf{MultiWOZ}} & \multicolumn{3}{c}{\textbf{SGD}}  \\ \hline
\% Seen Intents $\rightarrow$     & 25\%    & 50\%    & 75\%   & 25\%    & 50\%   & 75\%   & 25\%     & 50\%     & 75\%    & 25\%   & 50\%   & 75\%   \\ \hline
CT          & 0.1043  & 0.2055  & 0.2574 & 0.5018  & 0.7341 & 0.6542 & 0.2991   & 0.4371   & 0.6607  & 0.4523 & 0.5389 & 0.6160  \\
RZS         & 0.1214  & 0.1940   & 0.3207 & 0.6393  & 0.7727 & 0.7811 & 0.4566   & 0.4703   & 0.6951  & 0.6677 & 0.6578 & 0.6741 \\
Coach    & 0.1248  & 0.2258  & 0.3081 & 0.6070   & 0.7341 & 0.8104 & 0.4408   & 0.4505   & 0.6522  & 0.5888 & 0.6419 & 0.6725 \\ \hline
\ourmodel~\noiob & 0.1550   & 0.2631  & 0.4108 & 0.6495  & 0.9437 & 0.9378 & 0.5137   & 0.5529   & 0.7843  & 0.6861 & 0.7315 & 0.7704 \\
\ourmodel        & \textbf{0.1710}   & \textbf{0.2895}  & \textbf{0.4220}  & \textbf{0.8093}  & \textbf{0.9659} & \textbf{0.9764} & \textbf{0.5248}   & \textbf{0.5533}   & \textbf{0.8581}  & \textbf{0.7180}  & \textbf{0.7925} & \textbf{0.8324} \\
\bottomrule
\end{tabular}
\end{table*}

{\ourmodel} achieves its best performance on \myspecial{ATIS} dataset (see Table~\ref{tab:zs_atis}) as compared to other datasets. It highlights that zero-shot slot filling across different intents within a single domain is relatively easier than across domains, since \myspecial{ATIS} dataset consists of a single domain, i.e., airline travel.
On the contrary, \myspecial{SGD} dataset is the most comprehensive public dataset (i.e., it has 46 intents across 20 domains), yet our proposed method \ourmodel~has better performance on it (see Table~\ref{tab:zs_sgd}) than on \myspecial{SNIPS} and \myspecial{MultiWoz} datasets.
This calls attention to another critical point: dataset quality.
We observe that \myspecial{SGD} dataset is not only comprehensive but also has high quality semantic description for slot types and all the domains have enough training examples with minimal annotation errors (based on a manual study of a small stratified sample from the dataset).
For example, the slot types ``\myvalue{restaurant\_name}'', ``\myvalue{hotel\_name}'', and ``\myvalue{attraction\_name}'' belong to different domains, but are very similar to one another.
The rich semantic description of each slot type makes it easy for the model to transfer knowledge from one domain to new unseen domains with high F1 scores.
{\ourmodel} shows poor performance on \myspecial{SNIPS} dataset (see Table~\ref{tab:zs_snips}) as compared to other datasets, especially for intents: ``\myvalue{RateBook}'' and ``\myvalue{SearchScreeningEvent}''. This poor performance further highlights our previous point (i.e., quality of the dataset) since \myspecial{SNIPS} dataset does not provide any textual descriptions for slot types. Moreover, slot names (e.g., ``\myvalue{object\_name}'' and ``\myvalue{object\_type}'') convey very little semantic information, which exacerbates the challenge for the model to perform well for unseen domains. Finally, the results on \myspecial{MultiWOZ} dataset (see Table~\ref{tab:zs_multiwoz}) highlights that transferring knowledge to new unseen intents/domains is easier when some similar intent/domain is there in the training set. 
For example, the model is able to transfer knowledge for new unseen target intent ``\myvalue{Find} \myvalue{Hotel}'' (i.e., not in the training) from other similar intents such as, ``\myvalue{Find} \myvalue{Restaurant}'' and ``\myvalue{Book} \myvalue{Hotel}'' effectively. However, for the target domain ``\myvalue{Find} \myvalue{Attraction}'' that does not have any similar domain in the training set, the model shows relatively poor performance. Similar observations can also be made other competing models.

\begin{table*}[t!]
\footnotesize
\caption{F1 scores for all competing models where the model is trained on one dataset and tested on the rest. This setting resembles real-life scenarios.}
\label{tab:zs_per_dataset}
\begin{tabular}{l|ccc|ccc|ccc|ccc}
\toprule
\textbf{Method $\downarrow$} \textbf{Train Dataset $\rightarrow$}       & \multicolumn{3}{c|}{\textbf{SNIPS}} & \multicolumn{3}{c|}{\textbf{ATIS}}  & \multicolumn{3}{c|}{\textbf{MultiWOZ}} & \multicolumn{3}{c}{\textbf{SGD}}    \\ \hline
Test Dataset $\rightarrow$ & ATIS   & MultiWOZ & SGD    & SNIPS  & MultiWOZ & SGD    & SNIPS    & ATIS     & SGD     & SNIPS  & ATIS   & MultiWOZ \\ \hline
CT                  & 0.0874 & 0.1099   & 0.0845 & 0.0589 & 0.0725   & 0.0531 & 0.0646   & 0.0878   & 0.0616  & 0.1463 & 0.2290  & 0.1529   \\
RZS                 & 0.0915 & 0.1209   & 0.1048 & 0.0819 & 0.0809   & 0.0912 & 0.1496   & 0.2103   & 0.0875  & 0.1905 & 0.3435 & 0.2134   \\
Coach            & 0.1435 & 0.1191   & 0.1301 & 0.0976 & 0.0962   & 0.0871 & 0.1201   & 0.1730    & 0.1102  & 0.1795 & 0.3383 & 0.1903   \\ \hline
\ourmodel~\noiob         & 0.1544 & 0.1433   & 0.1504 & 0.1156 & 0.1124   & 0.1359 & 0.1242   & 0.1885   & 0.1258  & 0.2544 & 0.4714 & 0.2743  \\
\ourmodel                & \textbf{0.2080}  & \textbf{0.1832}   & \textbf{0.1690}  & \textbf{0.1436} & \textbf{0.1394}   & \textbf{0.1361} & \textbf{0.1847}   & \textbf{0.2662}   & \textbf{0.1620}   & \textbf{0.2761} & \textbf{0.5205} & \textbf{0.2884}   \\
\bottomrule
\end{tabular}
\end{table*}

\stitle{Comparison for seen and unseen slot types.}
An unseen target domains may have both unseen and seen slot types. The unseen ones have never been seen during training, and seen ones might have different contexts. For example, ``\myvalue{date}'' is a common slot type that may correspond to many different contexts in different domains such as date of a salon appointment, date of a restaurant booking, return date of a round-trip flight, and so on.
We evaluate the performance of the competing models on seen and unseen slot types individually to test each model's ability in handling completely unsen slot types.
Table~\ref{tab:seen_unseen} presents results in further detail where results for both seen and unseen slot types are reported separately.
{\ourmodel} is consistently better than other models on seen as well as unseen slot types.
On average, our proposed model {\ourmodel} shows 18\% and 17\% gains in F1 scores over the SOTA model for seen and unseen slots, respectively.
These gains are due to our slot-independent \myspecial{IOB} predictions (which provide effective templates for seen slot types) and our context-aware similarity function (which works well regardless whether slot types are seen or unseen).
Moreover, all the models have better performance on seen slots than on unseen ones as it is relatively easier to adapt to a new context (i.e., in new domain) for seen slots than to new unseen slots in an unseen context.
We also note that {\ourmodel} achieves a similar performance on \myspecial{ATIS} dataset for seen slots in the unseen target domain, when compared with the results reported by SOTA \emph{supervised slot filling} methods in~\cite{zhang2018joint}, i.e., F1 score of $0.952$ vs $0.959$ by our method.

\stitle{Train on a certain percentage of intents/domains and test on the rest.} 
Large labeled training datasets are an important factor in accelerating the progress of supervised models. To investigate whether zero-shot models are affected by the size of training data from different domains, we vary the size of the training data and report results to quantify the effect. Table~\ref{tab:zs_percent} presents results on all datasets when the training set has data from $\approx 25\%$, $\approx 50\%$, and $\approx 75\%$ of the intents/domains and the rest are used for testing.
The choice of intents/domains to be in the training or testing sets is done randomly, and average results are reported over five runs.
This setting is more challenging in two ways: models have access to less training data and the test utterances come from multiple domains.
{\ourmodel} is at least $19.06\%$ better (better F1 scores) than other models for any percentage of unseen intents on any dataset.
Overall, the performance of {\ourmodel} improves as it gets access to training data from more intents/domains, which is a desirable behaviour. Moreover, we also observe that our model achieves $0.72$ F1 score on \myspecial{SGD} with only 25\% of domains in the training data, which once again validates the intuition that having better quality data is very critical to adapt models to new unseen domains.
Similar results are observed on \myspecial{ATIS} dataset (i.e., single domain dataset), that highlights that knowledge transfer within a single domain is easier, and models can do a very good job on unseen intents even with a small amount of training data (e.g., 25\% intents in the training set). Similar conclusions hold true for other methods.

\begin{figure*}[ht]
\centering
\begin{subfigure}{.32\textwidth}
  \centering
  \includegraphics[width=\linewidth]{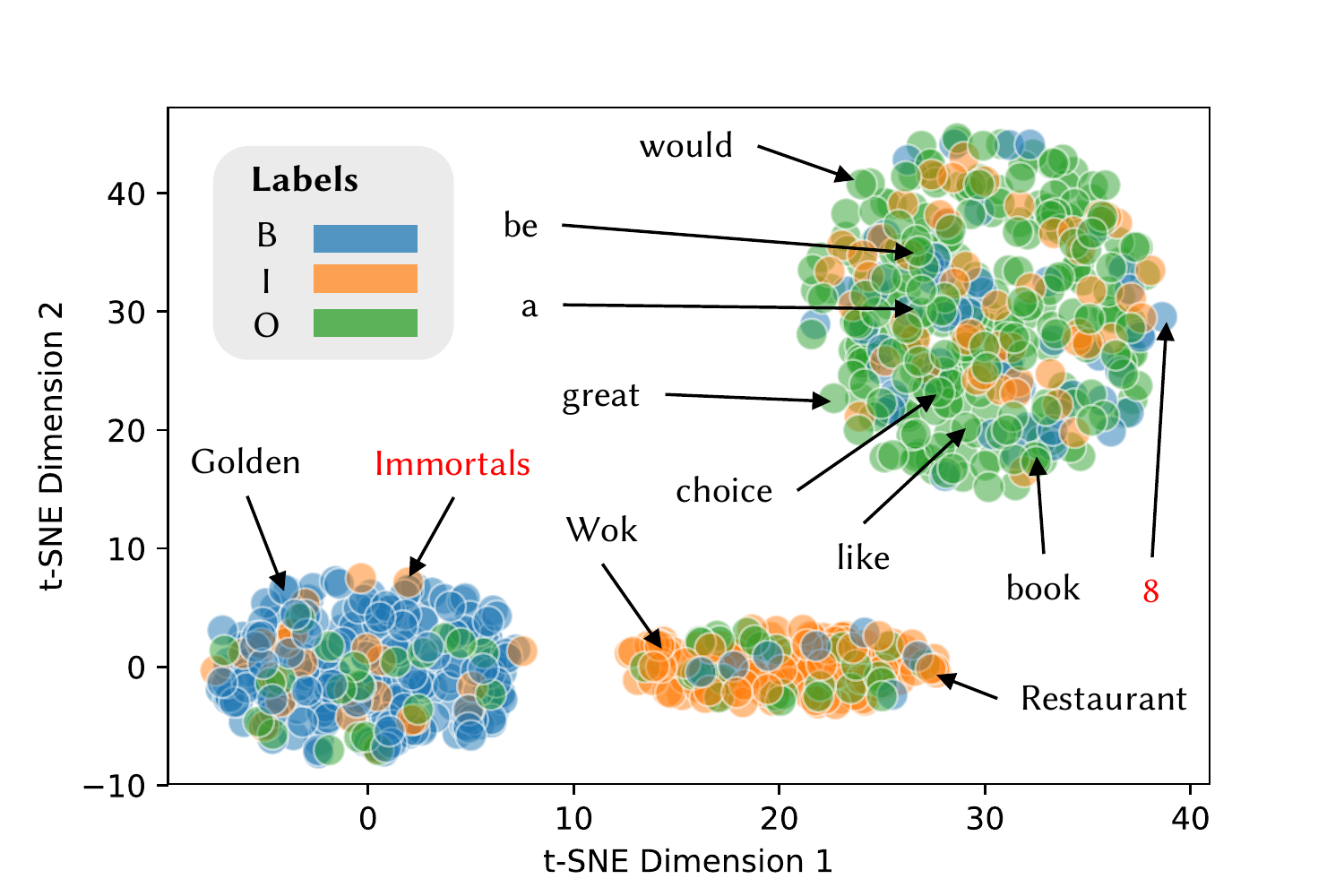}  
  \caption{RZS}
  \label{fig:viz-rzs}
\end{subfigure}
\begin{subfigure}{.32\textwidth}
  \centering
  \includegraphics[width=\linewidth]{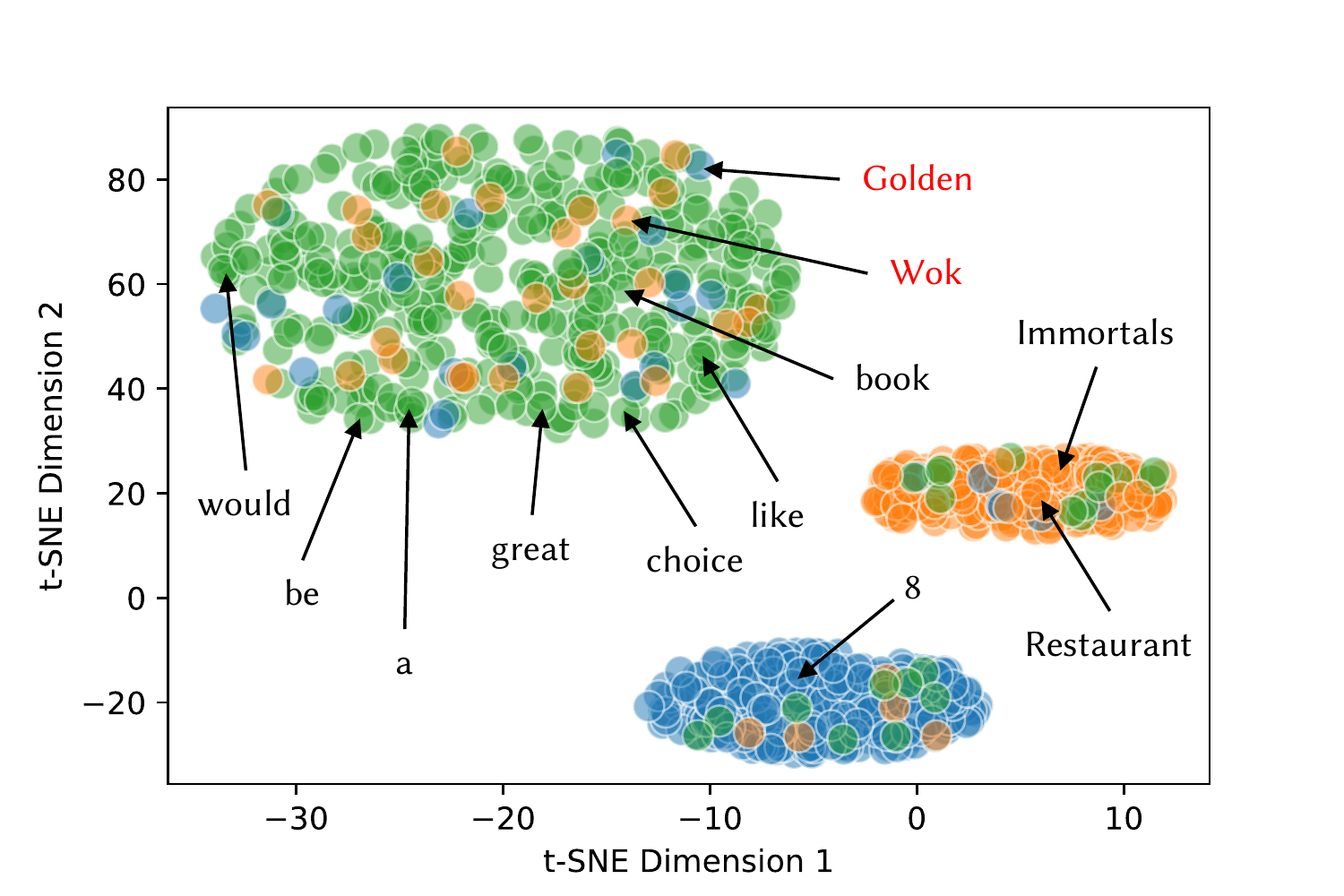}  
  \caption{Coach}
  \label{fig:viz-coach}
\end{subfigure}
\begin{subfigure}{.32\textwidth}
  \centering
  \includegraphics[width=\linewidth]{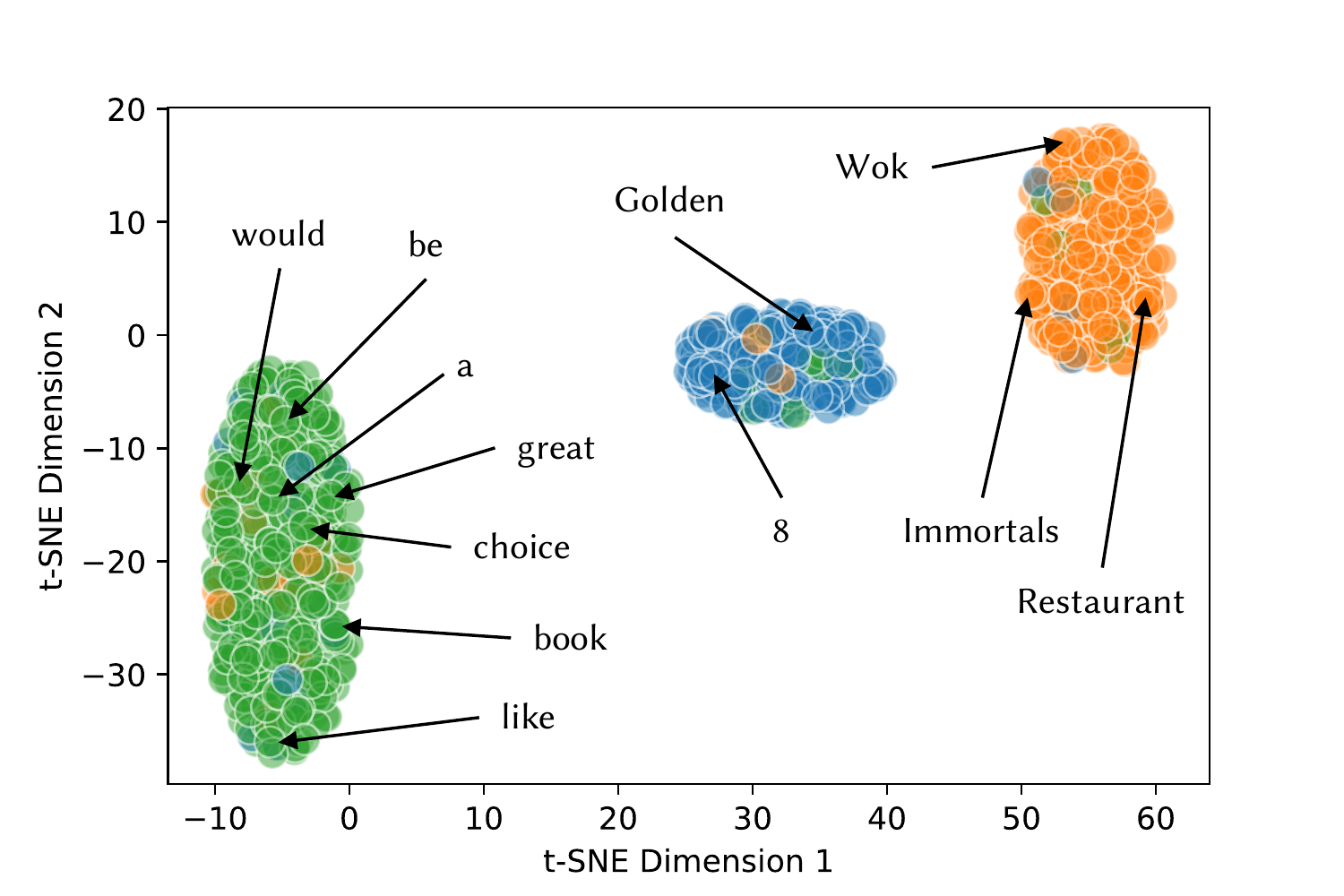}  
  \caption{LEONA (this work)}
  \label{fig:viz-leona}
\end{subfigure}
\caption{t-SNE visualization of word representations from selected utterances;  the selected utterances belong to the unseen domain ``Restaurant'' in SGD dataset and contain the slot type ``restaurant\_name''. Results are presented for the best performing 3 models.}
\label{fig:visualization-tsne}
\end{figure*}

\stitle{Train on one dataset and test on the rest of the datasets.}
This setting closely resembles the real-world zero-shot setting, where a model is trained on one dataset and tested on the rest.
This is the most challenging setting, since the test datasets come from purely different distributions than those seen during training. Although each domain within a dataset can be thought of as a different distribution, every dataset shows some similarity of expression across different domains. Table~\ref{tab:zs_per_dataset} presents results of all competing models for this setting. All models show relatively poor performance for this challenging setting. However, {\ourmodel} is consistently better than others; specifically, it is up to $56.26\%$ better on F1 score than the SOTA model. Our model achieves the best performance when it is trained on the \myspecial{SGD} dataset (relatively better quality dataset) and tested on the rest. On the contrary, it shows the worst performance, when trained on \myspecial{ATIS} (i.e., single-domain) and tested on the rest. Similar observations can be made for the other models. 
These results once again highlight the importance of the quality and comprehensiveness of the training dataset(s). 
Finally, this setting also indicates that the current SOTA models are not yet ready to be deployed in real-world scenarios and calls for more explorations and research in this important yet challenging and less-explored task of zero-shot slot filling.

\begin{table}[t!]
\caption{Ablation study of our model LEONA in the zero-shot setting: averaged F1 scores for unseen target domains.}
\label{tab:ablation}
\begin{tabular}{l|cccc}
\toprule
Configuration & \textbf{SNIPS}  & \textbf{ATIS}   & \textbf{MultiWOZ} & \textbf{SGD}    \\ \hline
Step 2        & 0.3689 & 0.6719 & 0.4792   & 0.6375 \\
Step 3        & 0.3812 & 0.6915 & 0.4999   & 0.6407 \\
Step 2 + 3    & 0.4013 & 0.7605 & 0.5412   & 0.6684 \\ \hline
Step 1 + 2    & 0.3820 & 0.6895 & 0.4936   & 0.6471 \\
Step 1 + 3    & 0.3866 & 0.7492 & 0.5594   & 0.6842 \\ \hline
Step 1 + 2 +3 & \textbf{0.4394} & \textbf{0.8397} & \textbf{0.5832}   & \textbf{0.7505} \\
\bottomrule
\end{tabular}
\vspace{-10pt}
\end{table}

\stitle{Ablation study.}
To quantify the role of each component in our model, we present an ablation study results in Table~\ref{tab:ablation} over all datasets.
First, we study the significance of the pre-trained NLP models in the first three rows in Table~\ref{tab:ablation}.
To produce the results in these rows, we used traditional word~\cite{bojanowski2017enriching} and character~\cite{hashimoto2016joint} embeddings instead of employing powerful pre-trained NLP models.
We observe that \myspecial{Step} \myspecial{three}, i.e., variant of the model that does not use pre-trained NLP models and does not consider ``\myvalue{IOB}'' tags from \myspecial{Step} \myspecial{two}, is the most influential component in the model, as it alone can outperform the best performing SOTA model Coach~\cite{liu2020coach}, but the margin is not significant (i.e., 0.3812 vs. 0.3739 on \myspecial{SNIPS}, 0.6915 vs. 0.6874 on \myspecial{ATIS}, 0.4999 vs. 0.4967 on \myspecial{MultiWOZ}, and 0.6407 vs. 0.6363 on \myspecial{SGD}).
If ``\myvalue{IOB}'' predictions from \myspecial{Step} \myspecial{two} are incorporated into it (i.e., row Step 2 + 3) or pre-trained NLP models are employed with it (i.e., row Step 1 + 3), its performance is further improved.
Moreover, if we just use \myspecial{Step} \myspecial{two} by predicting ``\myvalue{IOB}'' tags and assigning these ``\myvalue{IOB}'' tags to the slot type with the highest similarity (i.e., row Step 2), or combine \myspecial{Step} \myspecial{one} with \myspecial{Step} \myspecial{two} (i.e., row Step 1 + 2), we note that we do not achieve the best results.

\subsection{Qualitative Analysis}
\label{results-qualitative}
In this experiment, we randomly selected $100$ utterances in the unseen target domain ``Restaurant'' from the \myspecial{SGD} dataset and visually analyzed the performance of the competing models in extracting the values of the slot type ``\myvalue{restaurant\_name}'' from the selected utterances.
The goal of this experiment is to visually highlight the strengths/weaknesses of the competing models.
We retrieved the multi-dimensional numerical representations of the words in the selected utterances from the final layers of each model and reduced the number of dimensions of each representation to two using t-SNE~\cite{maaten2008visualizing}. Figure~\ref{fig:visualization-tsne} shows scatter plots for the resulting 2-dimensional representations for each model.
We observe that all models produce clear-cut clusters for each class:  \myvalue{B}, \myvalue{I}, or \myvalue{O}, which indicates that all models are able to produce distinguishing representations. However, {\ourmodel} produces better representations in the sense that less words are misclassified. That is, there are less violating data point in the clusters of {\ourmodel} in Figure~\ref{fig:visualization-tsne} (c).

We further analyze the results for two utterances: ``Golden Wok would be a great choice in ...'' and ``I would like to book a table at 8 Immortals Restaurant in ...''.
RZS~\cite{shah2019robust} is able to predict full slot value (i.e., \myvalue{Golden} \myvalue{Wok}) of the slot ``\myvalue{restaurant\_name}'' in the first utterance. However, we notice that RZS fails to capture the full value (i.e., ``\myvalue{8} \myvalue{Immortals} \myvalue{Restaurant}'') for the slot ``\myvalue{restaurant\_name}'' in the other utterance, where it could partially extract ``\myvalue{Immortals} \myvalue{Restaurant}'' and mistakenly assigns label \myvalue{O} to the word ``\myvalue{8}'', which led to subsequent wrong prediction for the word ``\myvalue{Immortals}'' (i.e., predicted label \myvalue{B}, whereas the true label is \myvalue{I}).
This misclassification is also highlighted in Figure~\ref{fig:visualization-tsne} (a) by coloring the wrongly predicted words with \myspecial{red}. Since RZS relies on the example value(s) and there is a high variability across the lengths of slot values, along with the diversity of expression, this model faces problems in detecting the \emph{full slot values}.

We notice that Coach~\cite{liu2020coach} fails to detect the value (i.e., \myvalue{Golden} \myvalue{Wok}) for the slot ``\myvalue{restaurant\_name}'' in the first utterance. However, it successfully captures the slot value in the other utterance. 
Since Coach relies on learning templates from seen domains and exploits those for unseen domains, it fails to handle the deviation of the unseen domains from the learned templates.
{\ourmodel} is able to detect full slot values for both utterances successfully, thanks to: the slot-independent \myspecial{IOB} predictions from \myspecial{Step} \myspecial{two}; the similarity function in \myspecial{Step} \myspecial{three} which is robust to errors from the previous steps; and the contextualization layers of the model.
Finally, we observed that our model also fails to fully detect very long slot values. For example, slot values 
``\myvalue{Rustic} \myvalue{House} \myvalue{Oyster} \myvalue{Bar} \myvalue{And} \myvalue{Grill}'',
``\myvalue{Tarla} \myvalue{Mediterranean} \myvalue{Bar} \myvalue{+} \myvalue{Grill}'', and
``\myvalue{Pura} \myvalue{Vida} \myvalue{-} \myvalue{Cocina} \myvalue{Latina} \myvalue{\&} \myvalue{Sangria} \myvalue{Bar}'' for the slot type ``\myvalue{restaurant\_name}'' are challenging to detect in unseen domains not only because of their long length, but also because of the presence of tokens like \myvalue{\&}, \myvalue{+}, and \myvalue{-}, that further exacerbate the challenge. 
Note that other SOTA models also fail to detect the above example slot values.
We plan to overcome this challenge in our future work by learning n-gram phrase-level representations to detect such slot values in their entirety. 

\section{Related Work}
\label{related}
We organize the related work into three categories: \myNum{i} supervised slot filling, \myNum{ii} few-shots slot filling, and \myNum{iii} zero-shot slot filling.

\stitle{Supervised Slot Filling.}
Slot filling is an extensively studied research problem in the supervised setting.
Recurrent neural networks such as Long Short-Term Memory (LSTM) or Gated Recurrent Unit (GRU) networks that learn how words within a sentence are related temporally~\cite{mesnil2014using,kurata2016leveraging} have been employed to tag the input for slots. Similarly, Conditional Random Fields (CRFs) have been integrated with LSTMs/GRUs~\cite{huang2015bidirectional,reimers2017optimal}.
The authors in~\cite{shen2017disan,tan2017deep} proposed a self-attention mechanism for sequential labeling. More recently, researchers have proposed jointly addressing the related tasks of intent detection and slot filling~\cite{goo2018slot, hakkani2016multi, liu2016attention, zhang2018joint, xu2013convolutional}. The authors in~\cite{zhang2018joint} suggested using a capsule neural network by dynamically routing and rerouting information from wordCaps to slotCaps and then to intentCaps to jointly model the tasks.
Supervised slot filling methods rely on the availability of large amounts of labeled training data from all domains to learn patterns of slot usage. In contrast, we focus on the more challenging as well as more practically relevant setting where new unseen domains are evolving and training data is not available for all domains.

\stitle{Few-shot Slot Filling.}
Few-shot learning requires a small amount of training data in the target domain. Meta-learning based methods~\cite{finn2017model, nichol2018reptile, nichol2018first} have shown tremendous success for few-shot learning in many tasks such as few-shot image generation~\cite{reed2017few}, image classification~\cite{snell2017prototypical}, and domain adaptation~\cite{vinyals2016matching}.
Following the success of such approaches, few-shot learning in NLP have been investigated for tasks such as text classification~\cite{sun2019hierarchical,geng2019induction,yan2018few}, entity-relation extraction~\cite{lv2019adapting,gao2020neural}, and few-shot slot filling~\cite{luo2018marrying,fritzler2019few,hou2020few}.
The authors in~\cite{luo2018marrying} exploited regular expressions for few-shot slot filling, Prototypical Network was employed in ~\cite{fritzler2019few}, and the authors in~\cite{hou2020few} extended the CRF model by introducing collapsed dependency transition to transfer label dependency patterns.
Moreover, few-shot slot filling and intent detection have been modeled jointly~\cite{krone2020learning,bhathiya2020meta}, where model agnostic meta learning (MAML) was leveraged. Few-shot slot filling not only requires a small amount of training data in the target domain, but also requires re-training/fine-tuning.
Our model addresses the task of zero-shot slot filling where no training example for the new unseen target domain is available and it can seamlessly adapt to new unseen domains -- a more challenging and realistic setting.

\stitle{Zero-shot Slot Filling.}
Zero-shot learning for slot filling is less explored, and only a handful of research has addressed this challenging problem, albeit, with very limited experimental evaluation.
Coach~\cite{liu2020coach} addressed the zero-shot slot filling task with a coarse-to-fine approach. It first predicts words that are slot values. Then, it assigns the predicted slot value to the appropriate slot type by matching the value with the representation of description of each slot type.
RZS~\cite{shah2019robust} utilizes example values of each slot type. It uses character and word embeddings of the utterance and slot types along with the slot examples' embeddings, and passes the concatenated information through a bidirectional LSTM network to get the prediction for each word in the utterance. 
CT~\cite{bapna2017towards} proposed LSTM network and employed slot descriptions to fill the slots for each slot type individually.
The authors in~\cite{lee2019zero} also employed LSTM, slot descriptions, and attention mechanisms for individual slot predictions. To tackle the challenge of the zero-shot slot filling, we leverage the power of the pre-trained NLP models, compute complex bi-directional relationships of utterance and slot types, and contextualize the multi-granular information to better accommodate unseen concepts.
In a related, but orthogonal line of research, the authors in~\cite{ma2019end,li2020sppd,gulyaev2020goal} tackled the problem of slot filling in the context of dialog state tracking where dialog state and history are available in addition to an input utterance. In contrast, this work and the SOTA models we compare against in our experiments only consider an utterance without having access to any dialog state elements.

\balance
\section{Conclusion}
\label{conclusion}
We have presented a zero-shot slot filling model, {\ourmodel}, that can adapt to new unseen domains seamlessly.
{\ourmodel} stands out as the first zero-shot slot filling model that effectively captures rich and context-aware linguistic features at different granularities.
Our experimental evaluation uses a comprehensive set of datasets and covers many challenging settings that stress models and expose their weaknesses (especially in more realistic settings).
Interestingly, our model outperforms all state-of-the-art models in all settings, over all datasets. The superior performance of our model is mainly attributed to: its effective use of pre-trained NLP models that provide domain-oblivious word representations, its multi-step approach where extra insight is propagated from one step to the next, its generalizable similarity function, and its contextualization of the words' representations.
In the most challenging evaluation setting where models are tested on a variety of datasets after being trained on data from one dataset only, our model is up to 56.26\% more accurate (in F1 score) than the best performing state-of-the-art model.
It remains challenging for all models, including ours, to identify slot values that are very long or that contain certain tokens. We plan to further improve our model by incorporating n-gram phrase-level representations to overcome this challenge and allow our model to accurately extract slot values regardless of their length or diversity.


\bibliographystyle{ACM-Reference-Format}
\bibliography{sample-base}
\end{document}